\documentclass[sigconf]{acmart}

\AtBeginDocument{%
  \providecommand\BibTeX{{%
    \normalfont B\kern-0.5em{\scshape i\kern-0.25em b}\kern-0.8em\TeX}}}

\renewcommand\footnotetextcopyrightpermission[1]{}

\acmConference[HabiTech, Workshop on Inhabiting Buildings, Data \& Technology, CHI'24]{ACM Conference}{May 11, 2024}{Honolulu, HI, USA}%

\begin{document}

\title{Research Challenges for Adaptive Architecture: Empowering Occupants of Multi-Occupancy Buildings}
\renewcommand{\shorttitle}{Research Challenges for Adaptive Architecture}

\author{Binh Vinh Duc Nguyen}
\email{alex.nguyen@kuleuven.be}
\orcid{0000-0001-5026-474X}
\affiliation{
  \institution{Research[x]Design, \break Department of Architecture, KU Leuven}
  \streetaddress{Kasteelpark Arenberg 1 - box 2431}
  \city{Leuven}
  \country{Belgium}
  \postcode{3001}
}

\author{Andrew Vande Moere}
\email{andrew.vandemoere@kuleuven.be}
\orcid{0000-0002-0085-4941}
\affiliation{
  \institution{Research[x]Design, \break Department of Architecture, KU Leuven}
  \streetaddress{Kasteelpark Arenberg 1 - box 2431}
  \city{Leuven}
  \country{Belgium}
  \postcode{3001}
}

\begin{abstract}
This positional paper outlines our vision of `adaptive architecture', which involves the integration of robotic technology to physically change an architectural space in supporting the changing needs of its occupants, in response to the CHI'24 workshop "\textit{HabiTech - Inhabiting Buildings, Data \& Technology}" call on "\textit{How do new technologies enable and empower the inhabitants of multi-occupancy buildings?}". Specifically, while adaptive architecture holds promise for enhancing occupant satisfaction, comfort, and overall health and well-being, there remains a range of research challenges of (1) how it can effectively support individual occupants, while (2) mediating the conflicting needs of collocated others, and (3) integrating meaningfully into the sociocultural characteristics of their building community. 
\end{abstract}


\begin{teaserfigure}
  \includegraphics[width=\textwidth]{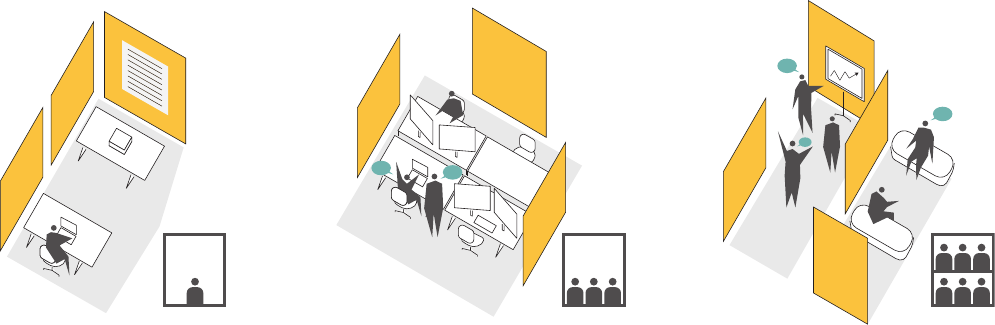}
  \caption{In this paper, we discuss how adaptive architecture should aim to address its remaining research challenges of (1) how it can effectively support each individual occupant, while (2) mediating the conflicting needs between collocated occupants, and (3) integrating meaningfully into the sociocultural characteristics of the overall building community.}
  \label{fig:teaser}
\end{teaserfigure}

\maketitle

In response to the new research area proposed by the CHI'24 workshop HabiTech - Inhabiting Buildings, Data \& Technology: "\textit{How do new technologies enable and empower the inhabitants of multi-occupancy buildings?}", this position paper presents our vision of \textit{adaptive architecture}, which involves the integration of robotic technology to physically change an architectural space. We discuss how this vision contributes to the proposed research area by making architectural spaces better support the changing needs of occupants and even `nudge' their behaviours towards health and wellbeing benefits, and outline its associated research challenges.

\section{Adaptive Architecture}
The surge in global population and urbanisation has increased the scarcity of space, which compelled architectural design to enhance its adaptability, i.e. allowing a single space to serve multiple functions asynchronously. This adaptability holds the potential to improve the sustainability and cost-efficiency of architecture, particularly when the layout of a single space can be flexibly adapted according to the changing - and often unpredictable - needs and activities of occupants. However, despite the commercial availability of furniture elements that allow an architectural layout to be adapted like movable partitions, rotating walls, sliding curtains or mobile furniture including partitions, cupboards, desks, chairs; their uptake into multi-occupancy buildings remains limited. This lack is primarily due to human constraints, such as the requirement to manually control or move these elements, which is physically taxing, time-consuming, socially embarrassing \cite{Obrien2014}, and often cannot be anticipated beforehand \cite{Zamani2019}.

\begin{figure*}[t]
    \centering
    \includegraphics[width=\linewidth]{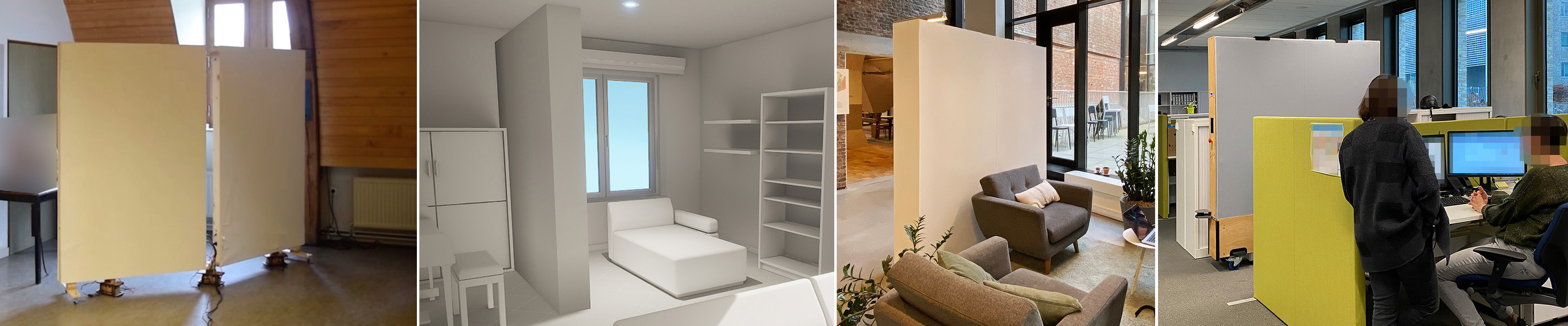}
    \caption{Our series of empirical studies with architectural-scale, semi-autonomous mobile robotic partitions that adapt the layouts \cite{Nguyen2020} of  multi-occupancy homes \cite{Nguyen2021} or offices \cite{Nguyen2022,Nguyen2024}.}
    \label{fig:deployment}
\end{figure*}

From the domain of architectural design, the field of adaptive architecture investigates how digital and robotic technology can automatically and autonomously adapt an architectural space to meet the changing needs of occupants. During five decades of research, visionaries in this field have proposed various theoretical manifestos \cite{Fox2016} and provocative installations \cite{Oosterhuis2008}, which demonstrate the potential of adaptive layout in not only hosting multiple functional purposes, but also evoking compelling architectural experience \cite{Achten2013, Meagher2015}. 
Thanks to recent scientific advancements of robotic furniture \cite{Ju2009, Lello2011, Sirkin2015} as well as the growing affordability of autonomous mobile robots capable of moving around layout elements like partitions \cite{Onishi2022}, sofas \cite{Spadafora2016}, chairs \cite{Agnihotri2019}, or tables \cite{Takashima2015}, this vision of adaptive layout has become much closer to reality. 
Evidently, a growing number of innovative start-ups like Ori\footnote{Ori: \href{https://www.oriliving.com/}{oriliving.com}}, Beyome\footnote{Beyome: \href{https://www.beyome.live/language/en/}{beyome.live}}, or Bumblebee\footnote{Bumblebee: \href{https://bumblebeespaces.com/}{bumblebeespaces.com}} are now offering robotic solutions to optimise space-use within apartments by flexibly moving large furniture items like beds or sofas between their usual locations and storage spaces nestled in underutilised areas, like beneath the floor, above ceilings, or inside space-delimiting cupboards.

\section{Sense of Place}
\begin{figure}
    \centering
    \includegraphics[width=0.7\linewidth]{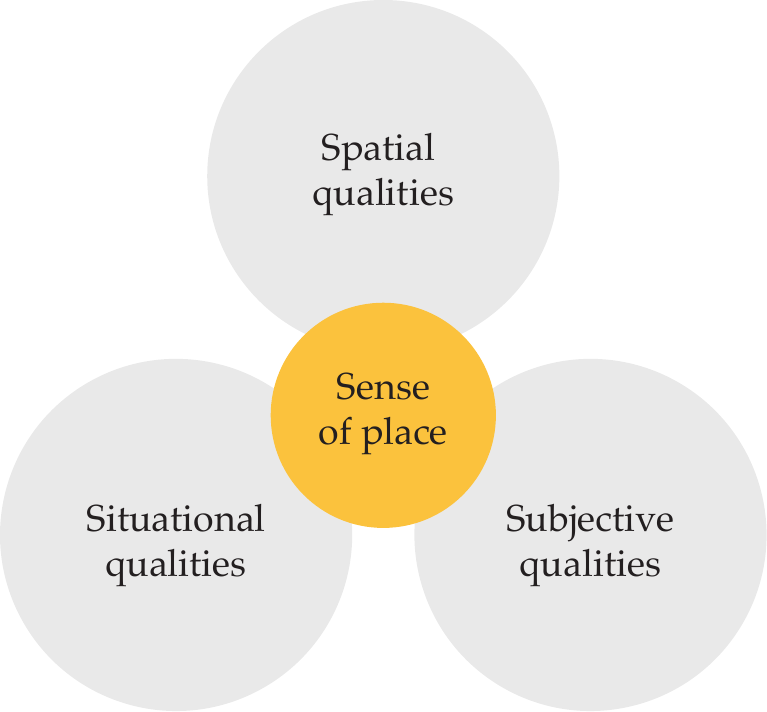}
    \caption{Our sense of place model, demonstrating how the occupant perception of a temporally-adapted layout hinges on the holistic alignment of its spatial, situational, and subjective qualities \cite{Nguyen2022}.}
    \label{fig:place}
\end{figure}

Motivated to explore the \textit{architectural impact} of these technologies beyond functional goals \cite{Wiberg2020}, our own research applied HCI methodology to capture how occupants actually experience adaptive layout in real-world contexts. 
Through a series of empirical studies with architectural-scale, semi-autonomous mobile robotic partitions that adapt the layouts \cite{Nguyen2020} of homes \cite{Nguyen2021} or offices \cite{Nguyen2022, Nguyen2024}, as shown in \autoref{fig:deployment}, we evidenced that occupants base their perception of a temporally-adapted layout on how it conveys an appropriate \textit{sense of place} \cite{Sime1986, Relph1976, Harrison1996}. This sense of place is experienced through the alignment of three qualities, as shown in \autoref{fig:place}, including:

\begin{itemize}
    \item \textit{Spatial qualities}: the physical manifestation of the adapted layout through the interplay of architectural elements.
    \item \textit{Situational qualities}: the temporal situation that surrounds the adapted layout, including both the social and the environmental context.
    \item \textit{Subjective qualities}: the personal lived experience or preferences of each individual occupant who experiences the adapted layout.
\end{itemize}

\noindent Our research showed that adaptive architecture should prioritise maintaining an appropriate sense of place. This entails unfolding a new adapted layout when one of the aforementioned qualities becomes misaligned (\textit{WHEN}), and doing so in a way that realigns the qualities to achieve an appropriate sense of place (\textit{HOW}). When this sense of place is ensured, occupants evidently not only accept adaptive architecture as appropriate, but also actively \cite{Nguyen2024} and purposefully \cite{Nguyen2022} utilise its adapted layouts, such as to enhance ergonomics, reduce disturbances, or even convey their intentions and negotiate their needs with others. Consequently, adaptive architecture has the potentials to improve the satisfaction and comfort of occupants, which have longer-term implications on their health, wellbeing \cite{Veitch2012, Scupelli2016}, and overall quality of life \cite{Steemers2020}.

However, despite these potentials, we also recognise that the practical implications of adaptive architecture in real-world multi-occupancy buildings would encounter challenges across multiple social levels. These challenges include addressing not only \textit{individual} concerns, i.e. how to capture and process the personal, and potentially sensitive, space-use data of occupants, but also \textit{group} dynamics, i.e. how to ensure the needs and preferences of multiple collocated occupants are appropriately considered and mediated, and \textit{community} engagement, i.e. how to meaningfully address the sociocultural characteristics of a building community.

\section{Research Challenges}
We thus detail our assessment on the aforementioned research challenges, and propose our vision for their mitigation.

\subsection{Capture individual experience}
To autonomously determine an appropriate adapted layout in a given situation, it is essential to continuously capture the three qualities contributing to the perceived sense of place. However, several research challenges persist regarding how this capturing process can unfold ethically and voluntarily yet robustly. For instance:

\begin{itemize}
    \item \textit{Accuracy}. The experience of an occupant on an adapted layout is closely tied to their nuanced perception of its spatial qualities, like the degree of openness to views, access to others, or exposure to natural light \cite{Nguyen2022}. Grasping these nuanced qualities requires precise capture of their space-use behaviours like location, posture, head orientation, or activities. Achieving such detailed capture in real-world buildings yet remains a challenge.
    \item \textit{Pervasiveness}. For adaptive architecture to respond effectively to the needs of an occupant, it must maintain awareness of situational qualities. However, since buildings comprise spaces with diverse privacy levels, and an occupant may perform various activities with differing privacy requirements, indiscriminate deployment of tracking technology should be avoided \cite{Schnadelbach2019}, which in turn could hinder the understanding of adaptive architecture on situations.
    \item \textit{Continuity}. The personal sensitivities and preferences of an individual occupant can evolve over time, whether it is due to long-term factors such as ageing or health changes, or short-term factors such as performing activities with different environmental requirements. Therefore, the understanding of adaptive architecture on subjective qualities must also be continuously monitored and updated.
\end{itemize}

\noindent Mitigation solutions can leverage advanced computer vision algorithms, which now allow for full anonymisation of occupants while still revealing their space-use behaviours through skeletonised representations \cite{Lee2019, Lee2021}. Digital twins simulation techniques \cite{Gnecco2023} can predict occupant spatial perception from limited data collected via pervasive sensing technologies \cite{Verma2017} or voluntary prompting devices \cite{Alavi2017}. Insights from human-robot interaction also indicate that occupants typically accept a robot in their everyday spaces only when they trust its good intentions \cite{Naneva2020} and believe in its helpfulness \cite{Lee2019Desk}. Similarly, occupants are generally willing to share their personal data only when they perceive the benefits of adaptive architecture \cite{Schnadelbach2019}. Therefore, perhaps adaptive architecture should aim to assess whether a correlation exists between its benefits and varying levels of data privacy, in order to enable occupants to make fully informed decisions regarding their privacy preferences.

\subsection{Mediate collocated impact}
An adapted layout inevitably influences its multiple collocated occupants in different and sometimes contrasting ways. Therefore, adaptive architecture must carefully consider its behaviours to mediate their diverse needs. Yet, it remains unclear how this collocated impact can be responsibly managed, such as:

\begin{itemize}
    \item \textit{Activity}. In a shared space, collocated occupants often simultaneously engage in activities that may conflict with one another, such as working versus playing in a home, or focusing versus collaborating in an office. As disturbances to one's activity are often caused by another \cite{Jackson2009}, manifesting an adapted layout to support an occupant may potentially hinder or even disrupt the activity of another.
    \item \textit{Equitability}. As multiple collocated occupants may choose to allow the collection of their individual space-use behaviour and experience at varying privacy levels, it is still uncertain how adaptive architecture should proceed to accommodate these preferences effectively, or whether it should proceed at all.
    \item \textit{Intentionality}. An adapted layout is visible to all collocated occupants, allowing them to employ it to communicate their intentions, such as signalling their availability (e.g. by hiding hiding behind a partition) \cite{Nguyen2022}, expressing their sensitivity (e.g. by shielding themselves from visual disturbances), or informing others about their needs (e.g. by moving a partition towards others to ask them to keep quiet) \cite{Nguyen2024}. Because an autonomously adapted layout may be perceived similarly, adaptive architecture should carefully manage the intentions it convey.
\end{itemize}    

\noindent While adaptive architecture can rely on occupants to actively control it and thus mediate their different needs by themselves, research in occupational health suggests that occupants often hesitate to take action for concern of negatively affecting others \cite{Dorn2013} or disrupting social dynamics \cite{Obrien2014}. Therefore, some level of autonomous behaviour is necessary. We thus propose that adaptive architecture should be able to identify situations when it should act autonomously, follow pre-programmed behaviors, or allow occupants full control. One potential mitigation approach is adopting the HRI autonomy framework of "shared control with robot initiative" \cite{Beer2014}, in which adaptive layout operates autonomously but pauses for input from occupants in uncertain scenarios. Another approach, inspired by previous studies on automated office service systems \cite{Ackerly2012, Lashina2019}, involves suggesting several possible layouts to occupants at appropriate moments and awaiting their decisions on implementation.

\subsection{Engage building community}
To appropriately integrate adaptive layout into a building, it is necessary to understand both the current sociocultural characteristics of its community, and how the integration of adaptive layout might motivate new sociocultural space-use patterns to emerge. As this process demands collaboration among not only occupants and building managers but also architects and and HCI designers, it has the potential to reshape the design, use, and management of buildings. For example:

\begin{itemize}
    \item \textit{Design}. Given that the perception of an appropriate sense of place, which defines how and when adaptive architecture should manifest, can vary among different sociocultural contexts, adaptive architecture must be designed by engaging the building community at hand. Furthermore, because the behaviour of adaptive architecture can then be fine-tuned or even re-designed, the role of architects should even transition from "providing static spaces" before occupancy to "orchestrating dynamic spaces" through continuous post-occupancy feedback. However, it remains uncertain how such evolving design process should unfold.
    \item \textit{Feedback}. Adaptive architecture empowers architectural space with the deliberate agency to autonomously adapt itself around the occupants and even `nudge' their space-use behaviours \cite{Onishi2022, Lee2013}. This agency has the potential to redefine how occupants `use' architectural spaces, such as by continuously interacting with it \cite{Oosterhuis2012} as an active agent \cite{Green2016} in a bi-directional dialog \cite{Schnadelbach2016} rather than a traditionally `static', passive backdrop. However, since occupants must retain the autonomy in determining when and to what extent their space-use behaviours are influenced, perhaps buildings should offer a spectrum of both adaptive and non-adaptive `static' spaces.
    \item \textit{Management}. Adaptive layout offers building managers the possibility to steer the space-use behaviours of occupants toward sustainability (e.g. `nudge' workers to utilise unoccupied spaces more efficiently to optimise office energy use) or health and well-being (e.g. `nudge' students to sit next to a window in a library to increase exposure to natural light) benefits. However, this process of `behavioural steering' could lead to tensions without clearly defined social rules and transparent communication of intentions, which in turn, require careful and effective design.
\end{itemize}

\noindent  As evidenced by prior research in smart buildings \cite{Rio2021} and adaptive facades \cite{Navarro2020}, enhancing the perceived agency of occupants over automatic building control systems cannot rely solely on technological advancements. It requires changes in building organisational culture, including effective collaborations and dialogues between building occupant communities and managers, and transparency in the collection and processing of building data \cite{Finnigan2020}. In the context of adaptive architecture, these cultural changes also necessitate the active involvement of architects and HCI designers, who possess the expertise to design meaningful adapted layouts with appropriate autonomous behaviours. Thus, it remains an open question whether adaptive architecture will foster a new kind of buildings that can be continuously monitored, evaluated, and improved for the benefits of their occupant communities.

\section{Conclusion}
This short positional paper presented our vision of `adaptive architecture' as an architectural-motivated approach to enable and empower building occupants by improving their satisfaction, comfort, and health and wellbeing. We identified its research challenges across three levels: individual, group, and community of building occupants, and proposed a potential research agenda to address these challenges.

\begin{acks}
This research is supported by the KU Leuven ID-N project IDN/22/003 "Adaptive Architecture: the Robotic Orchestration of a Healthy Workplace".
\end{acks}

\bibliographystyle{ACM-Reference-Format}
\balance
\bibliography{sample-base}


\end{document}